\documentclass{article}

\usepackage{microtype}
\usepackage{graphicx}
\usepackage{subcaption}
\usepackage{booktabs} 

\usepackage{hyperref}

\usepackage[preprint]{icml2026}

\usepackage{amsmath}
\usepackage{amssymb}
\usepackage{mathtools}
\usepackage{amsthm}

\usepackage[capitalize,noabbrev]{cleveref}

\theoremstyle{plain}

\theoremstyle{definition}

\theoremstyle{remark}

\usepackage[disable,textsize=tiny]{todonotes}

\icmltitlerunning{Sparse Inter-Layer Dependencies of Transformer FFN Neurons}

\begin{document}

\twocolumn[
  \icmltitle{Sparse Inter-Layer Dependencies of Transformer FFN Neurons}

  \icmlsetsymbol{equal}{*}

  \begin{icmlauthorlist}
    \icmlauthor{Johannes Knittel}{yyy}
    \icmlauthor{Hanspeter Pfister}{yyy}
  \end{icmlauthorlist}

  \icmlaffiliation{yyy}{Harvard University}

  \icmlcorrespondingauthor{Johannes Knittel}{jknittel@seas.harvard.edu}

  \icmlkeywords{Transformers, feedforward networks, sparsity, mechanistic interpretability, attribution}

  \vskip 0.3in
]

\printAffiliationsAndNotice{}  

\begin{abstract}

Feedforward network (FFN) blocks account for a large fraction of the parameters and computation in Transformer architectures, yet their internal structure remains difficult to interpret due to the additive superposition induced by the residual stream.
We examine whether the activation of an FFN neuron can be explained by a sparse set of preceding neuron activations and attention outputs.
We introduce a training-free attribution method that estimates the relative influence of upstream neurons and attention outputs on a target neuron's activation.
Empirically, across models and layers, we find that small subsets of preceding activations and attention outputs suffice to preserve neuron activations with high fidelity when all remaining inputs are masked with their average values.
Effective sparsity is even greater when accounting for the inherent activation sparsity of upstream layers.
Moreover, applying the neuron-specific masks in all layers simultaneously, such that the induced deviations propagate through the network, leaves model perplexity largely unchanged at moderate sparsity levels.
These results demonstrate that, despite dense parameterization, FFNs exhibit sparse and structured inter-layer dependencies at the neuron level.
Our method provides a practical, scalable tool for circuit-level interpretability and identifies candidate sparse pathways with potential implications for efficient inference.

\end{abstract}

\section{Introduction}
\label{sec:intro}

Feedforward network (FFN) blocks constitute the dominant parametric component of Transformer architectures~\citep{vaswaniAttentionAllYou2017}, typically accounting for two-thirds of model parameters, and they drive the majority of inference-time computation except for long sequences, where the quadratic complexity of the attention mechanism takes over.
Beyond their quantitative dominance, FFN blocks have been shown to serve as associative memories that store and retrieve factual knowledge and linguistic patterns~\citep{gevaTransformerFeedforwardLayers2021,gevaTransformerFeedforwardLayers2022}.
Yet despite their central role, the internal dynamics governing FFN activations remain poorly understood: given the complex, superposed inputs arriving via the residual stream, what determines whether a particular neuron fires?

The residual stream architecture poses a fundamental challenge to answering this question.
By design, every attention head and FFN block writes its output to a shared, high-dimensional representation through additive connections.
While essential for gradient flow and training stability, this architectural choice induces superposition~\citep{elhageToyModelsSuperposition2022}: the input to any FFN neuron is a dense linear combination of all preceding computations.
From the perspective of a single neuron, disentangling this mixture to identify the specific upstream components that trigger its activation appears intractable without additional structure.

Existing approaches to this challenge fall into two broad categories, each with significant limitations.
Mechanistic interpretability methods employ causal interventions such as activation patching~\citep{mengLocatingEditingFactual2022} to trace information flow, but their computational cost scales poorly and typically requires task-specific prompts, making neuron-level analysis across entire models impractical.
Alternatively, Sparse Autoencoders (SAEs)~\citep{brickenMonosemanticityDecomposingLanguage2023} learn disentangled feature bases from residual stream activations, revealing interpretable structure hidden within superposed representations.
However, SAEs require substantial training overhead and operate on learned features rather than the model's native neurons, leaving open the question of whether sparsity exists at the level of direct neuron-to-neuron connectivity.

In this paper, we ask whether sparse structure is intrinsic to FFN computation itself.
Specifically, we hypothesize that despite dense parameterization, each FFN neuron depends on only a small subset of upstream neurons and attention outputs, and that the seemingly monolithic connectivity of Transformers masks a sparse computation graph operating at the neuron level.
If true, this would suggest that the functional pathways of language models are highly selective, even without learned feature-extraction frameworks.

To test this hypothesis, we introduce a training-free attribution method that decomposes the pre-activation of any FFN neuron into contributions from all preceding components: upstream neuron outputs, attention outputs, and embeddings.
By combining weight-based alignment—measuring how strongly each upstream output vector projects onto a neuron's input weights—with the empirical variance of upstream activations, we derive relevance scores that rank the influence of every potential upstream contributor.
This decomposition requires no additional training, operates directly on model weights, and scales to all neurons in a model.

Our empirical evaluation across the GPT-2 family~\citep{radfordLanguageModelsAre2019} and Qwen2.5~\citep{yangQwen25TechnicalReport2024} reveals striking sparsity in inter-layer dependencies.
We find that small subsets of upstream components suffice to preserve FFN activation patterns with high fidelity under mean-ablation of all remaining components.
These sparse dependencies also support the computation of the model as a whole: when the neuron-specific masks are applied in all layers simultaneously, so that the induced deviations propagate through the entire network, perplexity remains close to that of the unmodified model at moderate sparsity levels.
Moreover, effective sparsity compounds when accounting for inherent activation sparsity: since only a fraction of upstream neurons are themselves active for any given input, the number of components that \emph{actively} influence a downstream neuron is smaller still.
We also observe structured patterns in dependency graphs, including locality (stronger dependence on adjacent layers) and systematic variation in layer-wise importance, with certain layers contributing more strongly than others.

These findings have several implications.
For mechanistic interpretability, sparse dependency graphs provide a practical starting point for circuit-level analysis, identifying candidate pathways at a complexity amenable to human inspection.
For architectural design, the observation that dense FFN layers operate via sparse functional pathways suggests opportunities for sparse-by-construction models that encode this structure directly.
Furthermore, the identified dependency patterns could guide the distillation of specialized models: by pruning or freezing components outside the sparse support for a given task or domain, one could derive compact architectures that exploit the underlying sparsity for faster inference without retraining from scratch.
More broadly, our results extend the evidence for linear representations in the residual stream to the computation itself: FFN neurons appear to operate on directions, reading from the superposed stream through the selective alignment of their input weights with the output directions of a few upstream components.
This view of Transformers as implementing selective, structured information flow may inform both interpretability research and efficiency-oriented model development.

\section{Related Work}

This work investigates the sparse inter-layer dependencies governing Transformer FFN activations.
Our research intersects with the functional role of FFNs, the theory of superposition, mechanistic circuit discovery, and attribution-based analysis of information flow.

\textbf{Functional Role of FFNs as Memory.} A foundational perspective treats FFN layers as key--value memories \citep{gevaTransformerFeedforwardLayers2021}, where input weights act as keys that retrieve and combine output-value vectors.
Subsequent work demonstrated that FFNs build predictions by promoting interpretable concepts into the residual stream \citep{gevaTransformerFeedforwardLayers2022}.
At the level of individual neurons, FFN units have further been linked to the expression of specific facts \citep{daiKnowledgeNeuronsPretrained2022} and to discrete features such as tokens, n-grams, and positions \citep{voitaNeuronsLargeLanguage2024}.
While these studies examine the \textit{semantics} of neurons, our work focuses on the \textit{mechanics of triggering}: identifying the sparse upstream components that act as the activation signals for these memories.

\textbf{Superposition vs. Native Neuron Sparsity.} The residual stream is often characterized as a communication channel under superposition, representing more features than its physical dimensionality \citep{elhageToyModelsSuperposition2022}.
To disentangle these features, Sparse Autoencoders (SAEs) have become a dominant tool for recovering monosemantic feature bases \citep{brickenMonosemanticityDecomposingLanguage2023, hubenSparseAutoencodersFind2024}, and transcoders extend this idea by learning sparsely activating approximations of entire FFN sublayers \citep{dunefskyTranscodersFindInterpretable2024}.
However, these approaches operate in learned latent spaces and require significant computational overhead for training.
Our work explores a more fundamental hypothesis: that sparsity is not just a property of hidden features, but a structural reality of the model's native neuron-to-neuron connectivity.
We shift the focus from discovering features to identifying the sparse "wiring" already present in the dense architecture.

\textbf{Mechanistic Circuit Discovery.} Mechanistic interpretability seeks to identify circuits, i.e., subgraphs of neurons and heads that perform specific tasks \citep{elhageMathematicalFrameworkTransformer2021, cammarataThreadCircuits2020}.
Causal methods like activation patching \citep{mengLocatingEditingFactual2022} and path patching \citep{wangInterpretabilityWildCircuit2023} provide rigorous evidence for these circuits but are often too computationally intensive for global, model-wide analysis.
Gradient-based approximations substantially reduce this cost \citep{kramarAtPEfficientScalable2024}, and automated discovery methods scale circuit identification to entire behaviors, operating on model components \citep{conmyAutomatedCircuitDiscovery2023} or learned SAE features \citep{marks2025sparse}; nevertheless, these approaches remain tied to task- or behavior-specific prompts.
Our training-free attribution method provides a scalable, global characterization of dependencies without requiring specific behavioral probes.

\textbf{Linear Representation Hypothesis.} Our analysis is informed by the \emph{Linear Representation Hypothesis} (LRH), which posits that many semantic or functional properties of a model correspond to approximately linear directions in activation space \citep{mikolov2013efficient, alain2017understanding, parkLinearRepresentationHypothesis2024,knittel2024gpt}.
This hypothesis underlies linear probing~\citep{alain2017understanding}, activation steering~\citep{subramaniExtractingLatentSteering2022}, and recent work formalizing concept intervention via causal inner products \citep{parkLinearRepresentationHypothesis2024}.
Rather than testing whether concepts themselves are linearly encoded, we adopt a structural perspective: if meaningful information is linearly accessible, then the mechanisms that gate and transmit this information (such as FFN input weights) should exhibit selective linear alignment with upstream components.
By decomposing FFN pre-activations into weight-aligned contributions, we study whether FFN keys are sparsely aligned with the values produced by earlier layers, providing a mechanistic extension of the LRH from representational geometry to inter-layer connectivity.

\textbf{Attribution and Sparse Computation Graphs.} Our approach relates to attribution frameworks such as Integrated Gradients \citep{sundararajanAxiomaticAttributionDeep2017} and Layer-wise Relevance Propagation (LRP) \citep{shrikumarLearningImportantFeatures2017, achtibatAttnLRPAttentionawareLayerwise2024}.
\citet{ferrandoInformationFlowRoutes2024} use attribution to view Transformers as sparse information-flow graphs during prediction.
Similarly, \citet{gevaDissectingRecallFactual2023} trace how information is refined across layers via "logic chains."
We extend this line of inquiry by providing a systematic analysis of \textit{inter-layer dependency sparsity} across all neurons.
Unlike methods that focus on end-to-end token predictions, we isolate the internal connectivity of FFN blocks, demonstrating that even "dense" layers operate via highly selective, sparse pathways.

\textbf{Activation and Structural Sparsity.} That neural activations are sparse is a well-documented phenomenon, particularly in ReLU-based models \citep{nairRectifiedLinearUnits2010, glorotDeepSparseRectifier2011}.
In Transformers, sparse activations have been leveraged for inference efficiency \citep{liLazyNeuronPhenomenon2023, voitaAnalyzingMultiheadSelfattention2019, mirzadehReLUStrikesBack2023}, for instance by predicting input-dependent subsets of attention heads and FFN neurons on the fly \citep{liuDejaVuContextual2023} or by regrouping FFN neurons into sparsely activated experts \citep{zhangMoEficationTransformerFeedforward2022}.
Furthermore, the Lottery Ticket Hypothesis~\citep{frankleLotteryTicketHypothesis2019} and weight pruning studies~\citep{DBLP:journals/corr/HanMD15} suggest that dense networks contain highly effective sparse subnetworks.
Our work bridges these two concepts, activation sparsity and structural pruning, by showing that FFN neurons are not only sparsely active but are also sparsely \textit{triggered} by their predecessors.

\section{Problem Formulation and Notation}
\label{sec:notation}

Transformer architectures interleave multi-head self-attention with position-wise feedforward network (FFN) blocks, both operating on a shared residual stream.
Given an input sequence of hidden states $X \in \mathbb{R}^{T \times d_{\text{model}}}$, each Transformer layer applies these blocks independently at each token position, while preserving information through residual connections.

In GPT-like pre-normalization Transformers, the feedforward block takes the form
\begin{equation}
\label{eq:ffn_block}
\mathrm{FFN}(X) = X + W_2 \, \sigma\!\left(W_1 \, \mathrm{LN}(X)\right),
\end{equation}
where $\mathrm{LN}(\cdot)$ denotes layer normalization, $W_1 \in \mathbb{R}^{d_{\text{ff}} \times d_{\text{model}}}$ and $W_2 \in \mathbb{R}^{d_{\text{model}} \times d_{\text{ff}}}$ are learned weight matrices, and $\sigma(\cdot)$ is an elementwise nonlinearity such as GELU.
The intermediate dimensionality $d_{\text{ff}}$ is typically several times larger than $d_{\text{model}}$, causing FFN blocks to dominate the parameter count of the model.
LayerNorm plays a central role by centering and normalizing the residual stream prior to the linear transformation.

RMSNorm can be seen as a simplification of this step that omits centering.

The FFN can be viewed as a collection of $d_{\text{ff}}$ neurons, each computing an activation
\begin{equation}
\label{eq:ffn_neuron}
a_i = \sigma\!\left(w_i^\top \mathrm{LN}(x_t)\right),
\end{equation}
where $w_i^\top$ denotes the $i$-th row of $W_1$.
These activations are linearly combined and added to the residual stream via $W_2$.
The magnitude of $a_i$ determines the contribution of the corresponding column ${W_2}_i$ to the residual update.
Under the key--value memory interpretation of FFNs~\citep{gevaTransformerFeedforwardLayers2021}, each row $w_i$ of $W_1$ acts as a key that reads from the residual stream, detecting how strongly the current representation matches a learned pattern, and the resulting activation $a_i$ determines how strongly the associated concept direction, the column ${W_2}_i$, is written back.
The FFN block as a whole can thus be seen as retrieving and linearly combining the concept vectors whose keys match the input.
Since the terms key and value are already reserved for the projections of the attention mechanism, we refer to ${W_2}_i$ as the \emph{output vector} of neuron $i$.
Due to the additive residual connection, FFN outputs are superposed with the incoming representation, complicating direct attribution of downstream behavior to individual FFN neurons.

A similar reading--writing interpretation applies to the attention block, although it should not be conflated with the Query--Key--Value projections of the attention mechanism itself.
For each token, the attention weights average the Value-projected representations of the attended positions, yielding, after concatenation across heads, an aggregated vector $v \in \mathbb{R}^{d_{\text{model}}}$ that is mapped to the residual stream by the output projection matrix $W_O$.
Analogous to FFN neurons, each dimension $v_i$ of this aggregated representation acts as a scalar activation that determines how strongly the corresponding column ${W_O}_i$ is added to the residual stream.
In this view, the attention weights and the Value projection govern which signals are read, while the columns of $W_O$ define the concept directions written back; the key difference from FFNs is that reading spans the representations of multiple token positions rather than the current position alone.
We refer to $v_i$ as the \emph{attention activation} of attention dimension $i \in \{1, \dots, d_{\text{model}}\}$, and call ${W_O}_i$ the \emph{output vector} of attention dimension $i$.
As in the FFN case, the residual addition entangles these contributions with the existing representation, complicating the attribution of model behavior to individual attention components.

In this paper, we investigate the connectivity between neural activations within a given layer and both neural and attention activations in preceding layers.
More specifically, we examine whether the activation $a_i^{(l)}$ of neuron $i$ in layer $l$ is determined by a sparse subset of neural activations and attention activations from earlier layers.

\section{Sparse Subset Analysis of FFNs}

To investigate the sparsity of neural dependencies on earlier neurons and attention activations, we decompose FFN pre-activations into additive contributions from embeddings, attention outputs, and neuron outputs, define a relevance score that quantifies the influence of these components, and causally test, via mean-ablation activation patching, whether a small subset of high-relevance components suffices to determine FFN activations.

\subsection{Decomposing Neural Activations}

The input to the FFN block at layer $l$ is given by the residual stream, which aggregates multiple additive contributions.
For token position $t$, we write the residual stream as
\begin{equation}
\label{eq:residual_decomposition}
x_t^{(l)} = \sum_{s \in \mathcal{S}^{(l)}} c_s \, u_s,
\end{equation}
where $\mathcal{S}^{(l)}$ indexes all contributing components up to layer $l$, including the token embedding and positional embedding as well as FFN neuron outputs, attention outputs, and biases from preceding layers.
Each $u_s \in \mathbb{R}^{d_{\text{model}}}$ denotes a centered output vector (i.e., $\mathbf{1}^\top u_s = 0$), and $c_s \in \mathbb{R}$ is the corresponding scalar activation: the neuron activation $a_i$ for FFN outputs, the attention activation $v_i$ for attention outputs, and a constant of one for embeddings and bias terms.
Since the centering performed by layer normalization is linear, it distributes over the sum, allowing us to work directly with centered output vectors.

We absorb the affine transformation induced by layer normalization into the parameters of the subsequent layer and treat the normalization step itself as centering and normalization only, following \citet{elhageMathematicalFrameworkTransformer2021}. Under this convention, the pre-activation of neuron $i$ in layer $l$ can be written as
\begin{equation}
\label{eq:preactivation}
z_i^{(l)} \;=\; \tilde{w}_i^{(l)\top} \, \hat{x}_t^{(l)} + \tilde{b}_i^{(l)},
\end{equation}
where $\hat{x}_t^{(l)}$ denotes the normalized residual stream, $\tilde{w}_i^{(l)}$ is the effective input weight vector after absorbing the layer normalization parameters, and $\tilde{b}_i^{(l)}$ is the corresponding effective bias.
This bias term subsumes both the original FFN bias and the shift introduced by layer normalization.

Substituting Eq.~\eqref{eq:residual_decomposition} into Eq.~\eqref{eq:preactivation} and using the linearity of the dot product yields
\begin{equation}
\label{eq:preactivation_decomposition}
z_i^{(l)} \;=\;
\frac{
\sum_{s \in \mathcal{S}^{(l)}} c_s \, \langle \tilde{w}_i^{(l)}, u_s \rangle
}{
Z
}
\;+\; \tilde{b}_i^{(l)},
\end{equation}
where $Z$ denotes the normalization denominator induced by layer normalization, given by the standard deviation of the pre-normalized residual stream,
\[
Z \;=\; \sqrt{\tfrac{1}{d}\sum_{j=1}^d \big(x^{(l)}_{t,j} - \mu^{(l)}_t\big)^2 + \epsilon},
\]
with residual stream dimension $d$ and mean $\mu^{(l)}_t$; we suppress the dependence of $Z$ on the token position and layer for readability.

This decomposition formalizes the FFN pre-activation as an aggregate of individual component-to-neuron interactions from all preceding neural and attention activations, along with the initial token embedding, positional embedding, and bias terms.
In a geometric view~\citep{parkLinearRepresentationHypothesis2024}, the FFN neuron acts as a linear probe that selectively filters the superposed information within the residual stream.
Each contribution depends on both the activation magnitude $c_s$ and the alignment between the neuron's input weights and the corresponding centered output vector, as quantified by their dot product.
Large pre-activations arise when highly active upstream components are well aligned with the neuron's input weights, whereas components with negative alignment inhibit activation.

\subsection{Relevance Scoring}

An important empirical observation is that the majority of dot products between neuron input weights and output vectors from preceding layers are close to zero.
This sparsity can be exploited to identify a small subset of neurons and attention dimensions that are potentially most relevant for a given target neuron.
Since these dot products depend solely on the learned weights, they can be pre-computed independently of any specific input and serve as a first, architecture-level indicator of potential influence.
While the dot product defines the architectural alignment between components, it does not account for the dynamic intensity of these signals during inference.
To bridge the gap between static weight geometry and functional importance, we adopt a variance-based relevance score: since the pre-activation in Eq.~\eqref{eq:preactivation_decomposition} is a sum of per-component contributions, we quantify the influence of a component on the target neuron by the variance of its contribution across inputs.

Let $\hat{c}_s = c_s / Z$ denote the LayerNorm-normalized activation of component $s$, so that its contribution to the pre-activation of neuron $i$ in layer $l$ is $\hat{c}_s \, \langle \tilde{w}_i^{(l)}, u_s \rangle$, where $u_s$ is the component's centered output vector and $\tilde{w}_i^{(l)}$ the effective input weight vector of the target neuron.
Because the dot product is a constant factor for a given component--neuron pair, the variance of this contribution factorizes as
\begin{equation}
\label{eq:final_relevance_score}
\mathcal{R}_{s \rightarrow i}^{(l)}
=
\mathrm{Var}\!\left( \hat{c}_s \, \langle \tilde{w}_i^{(l)}, u_s \rangle \right)
=
\langle \tilde{w}_i^{(l)}, u_s \rangle^2 \, \mathrm{Var}(\hat{c}_s).
\end{equation}
The activation variance $\mathrm{Var}(\hat{c}_s)$ is independent of any particular target neuron and can be estimated once per component from sample activations, so scoring all component--neuron pairs reduces to multiplying the squared dot product by a pre-computed scalar.
Unlike threshold-based notions of activity, the variance treats strongly positive and strongly negative deviations symmetrically, which is important for attention dimensions whose contributions may take either sign.

The resulting ranking is thus determined by two factors: the magnitude of the dot product, capturing the geometric alignment between the component's output vector and the neuron's input weights, and the standard deviation of the normalized activation, capturing jointly how strongly and how often the component activates.
Components that are either geometrically orthogonal to the target neuron ($\langle \tilde{w}_i, u_s \rangle \approx 0$) or whose activations rarely deviate from their mean ($\mathrm{Var}(\hat{c}_s) \approx 0$) are assigned low relevance.
For token and positional embeddings, we consider only their alignment with the input weights, as their contributions to the residual stream are either present with unit activation or absent.
This score allows us to rank the communication channels within the residual stream, identifying the sparse subset of upstream pathways that effectively dominate the activation of a given FFN neuron.

\subsection{Causal Analysis via Activation Patching}
\label{sec:sparseReconstruction}

\begin{figure*}[htbp]
  \begin{center}
    \centerline{\includegraphics[width=\linewidth]{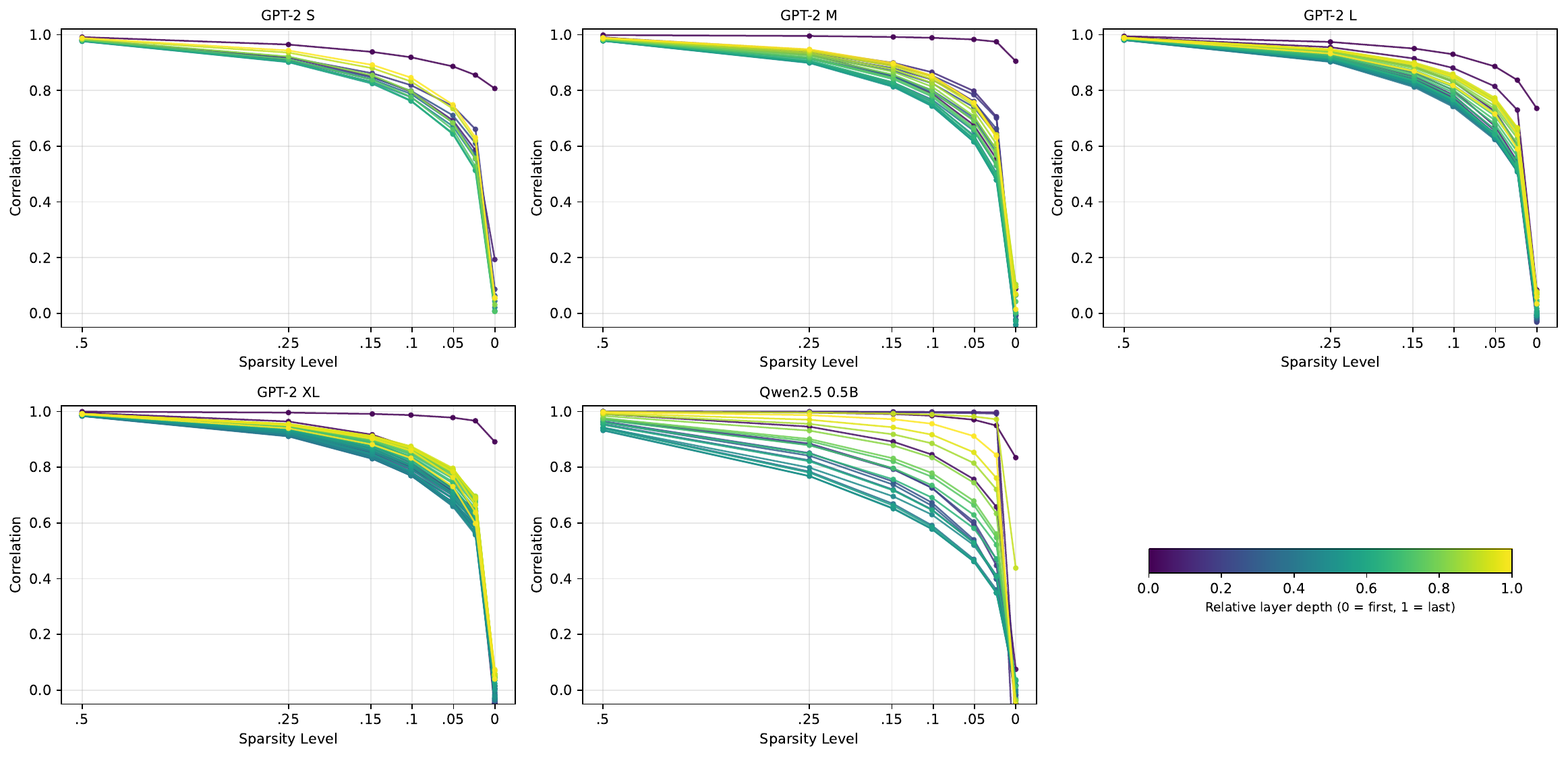}}
    \caption{
Layer-wise fidelity of FFN neuron pre-activations when all upstream components (neurons and attention dimensions) outside the selected sparse subset are mean-ablated, as a function of the sparsity level (the fraction of retained upstream components). Values show the average Pearson correlation between actual and patched pre-activations.}

    \label{fig:layerCorrPlot}
  \end{center}
\end{figure*}

\begin{figure*}[htbp]
  \begin{center}
    \centerline{\includegraphics[width=\linewidth]{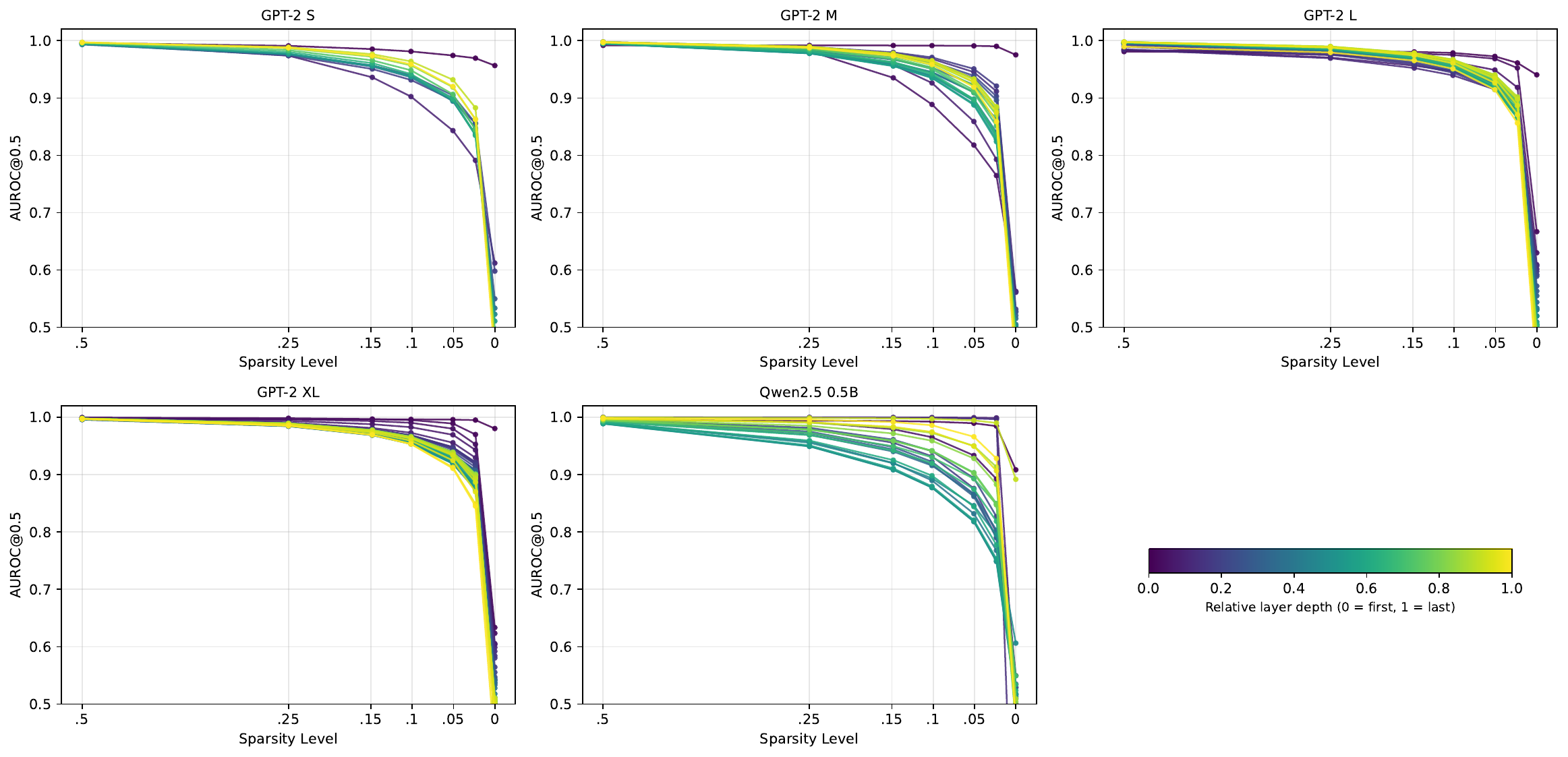}}
    \caption{
Layer-wise fidelity of FFN neuron pre-activations under the same intervention as in Figure~\ref{fig:layerCorrPlot}. Values show the AUROC for predicting meaningfully active neurons from patched pre-activations.}

    \label{fig:layerAurocPlot}
  \end{center}
\end{figure*}

To assess whether the highest-ranked components indeed determine a neuron's activation, we perform a causal intervention in the style of activation patching with mean ablation.
Concretely, for a target neuron $i$ in layer $l$, we select a subset $\mathcal{S}^{(l)}_K \subset \mathcal{S}^{(l)}$ consisting of the top-$K$ neurons and attention dimensions ranked by $\mathcal{R}_{s \rightarrow i}^{(l)}$, together with the token embedding and, where applicable, the positional embedding.
We then mask the direct contributions of all remaining components by replacing their activations with the empirical mean $\bar{c}_s = \mathbb{E}[c_s]$, while the selected components retain their actual activations.
The patched pre-activation is given by
\begin{equation}
\label{eq:patched_preactivation}
\check{z}_i^{(l)} =
\frac{
\sum_{s \in \mathcal{S}^{(l)}_K} c_s \, \langle \tilde{w}_i^{(l)}, u_s \rangle
+
\sum_{s \notin \mathcal{S}^{(l)}_K} \bar{c}_s \, \langle \tilde{w}_i^{(l)}, u_s \rangle
}{
Z
} \;+\; \tilde{b}_i^{(l)}.
\end{equation}
The mask is specific to each target neuron and applies only to the direct contributions entering the neuron's dot product; the intervention is not propagated through the network.
The normalization denominator $Z$ is taken from the actual, unpatched forward pass, so the intervention affects only the component activations, and the mean activations $\bar{c}_s$ are pre-computed from sample data.
If the patched pre-activations closely track the actual ones, the selected sparse subset suffices to determine the neuron's activation, with all remaining direct inputs reduced to their average effect.

\section{Experiments}

\begin{table}[t]
\caption{Model variants and core architectural parameters.}
\label{tab:models}
\centering
\setlength{\tabcolsep}{4pt}
\renewcommand{\arraystretch}{1.1}
\begin{tabular}{lcccc}
\hline
Model & Layers & $d_{\text{model}}$ & $d_{\text{ff}}$ & Params \\
\hline
GPT-2 S  & 12 & 768  & 3072 & 124M \\
GPT-2 M  & 24 & 1024 & 4096 & 355M \\
GPT-2 L  & 36 & 1280 & 5120 & 774M \\
GPT-2 XL & 48 & 1600 & 6400 & 1.5B \\
\hline
Qwen2.5-0.5B & 24 & 896  & 4864 & 490M \\
\hline
\end{tabular}
\end{table}

\subsection{Setup}

We evaluate our sparse activation-patching analysis across several GPT-2~\citep{radfordLanguageModelsAre2019} variants, as well as on Qwen2.5-0.5B~\citep{yangQwen25TechnicalReport2024}, a more recent architecture incorporating rotary embeddings, RMSNorm, and the SiLU activation function.
Table~\ref{tab:models} summarizes the models and their key architectural parameters.
For each model, we apply our method to all FFN neurons across all layers and analyze the fidelity of patched pre-activations under varying sparsity levels.

To compute relevance scores and the average activation values used for masking, we rely on empirical activation statistics collected from a subset of the MiniPile~\citep{kaddourMiniPileChallengeDataEfficient2023} training dataset.
These statistics are used exclusively to estimate the activation variances and means of upstream components.
To mitigate overfitting to specific activation patterns, all patching experiments and evaluations are performed on a disjoint subset of the MiniPile test dataset.
The MiniPile dataset is a diverse, deduplicated subset of the Pile~\citep{gaoPile800GBDataset2020}, drawing from a wide range of sources that span multiple domains and language styles.

For a given neuron and sparsity level, we compute its patched pre-activation with the neuron-specific mask in place, as described in Section~\ref{sec:sparseReconstruction}.
We then evaluate the fidelity of the intervention using two complementary metrics.
First, we compute the Pearson correlation between the actual and patched pre-activation values of each neuron across evaluation tokens, and report the average over all neurons of a layer.
Second, we report the area under the receiver operating characteristic curve (AUROC) for predicting \emph{meaningfully active} neurons, defined by a post-activation value greater than $0.5$, from the patched pre-activations.
The AUROC contextualizes the absolute deviations captured by the correlation, measuring whether the intervention preserves the distinction between active and inactive neurons.

\subsection{Main Results}
\label{sec:results}

\begin{figure*}[tb]
  \begin{center}
    \centerline{\includegraphics[width=\linewidth]{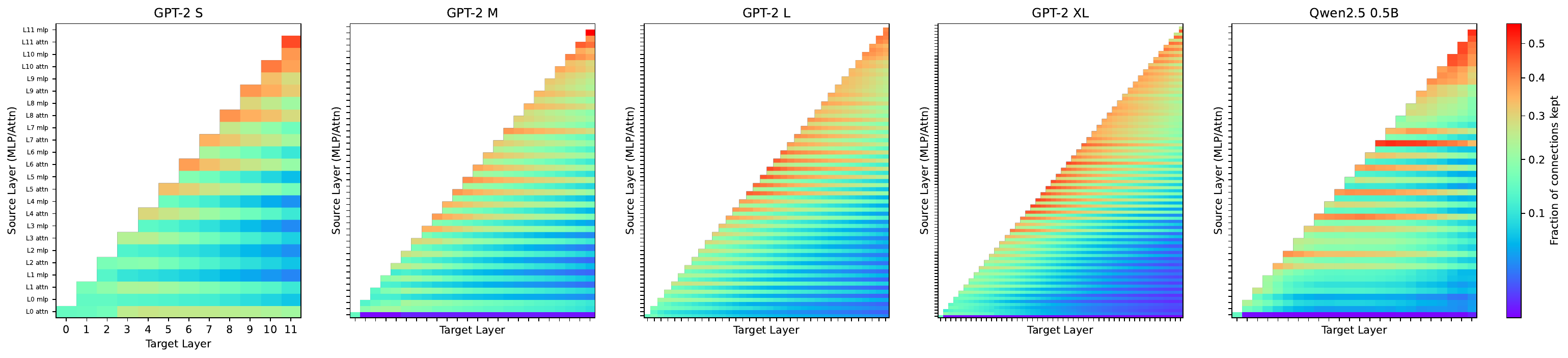}}
    \caption{
Layer-wise origin of selected upstream components (neurons and attention dimensions; y-axis) for each target layer (x-axis), computed at a sparsity level of $15\%$. Values indicate the fraction of each source layer's components that are retained in the selected subsets, averaged over the target layer's neurons.
}

    \label{fig:layerDensityPlot}
  \end{center}
\end{figure*}

Figures~\ref{fig:layerCorrPlot} and~\ref{fig:layerAurocPlot} summarize the fidelity of patched pre-activations across layers and model sizes, where the sparsity level denotes the fraction of upstream components retained.
Across all evaluated models, we observe consistently strong correlations and high AUROC scores, particularly at sparsity levels of $10\%$ and above.
Even at 5\% sparsity, most layers maintain AUROC scores above 0.9, indicating that the selected sparse subsets of upstream components suffice to preserve the activation patterns of FFN neurons with high fidelity.
Fidelity varies more strongly across layers in Qwen2.5-0.5B than in the GPT-2 family, with early and late layers reaching notably higher fidelity than middle layers.
These results suggest that a relatively small fraction of upstream neurons and attention dimensions suffices to determine FFN activations, even when the direct contributions of all remaining components are reduced to their average effect, consistent with substantial functional sparsity in inter-layer dependencies.

This degree of sparsity is somewhat unexpected in light of the common characterization of the residual stream as a highly superposed representation, where information from many features is distributed across dimensions.
Under such a view, one might anticipate FFN neurons to depend on more broadly distributed upstream signals.

The reported sparsity levels should be interpreted as upper bounds on the number of upstream components required to determine a neuron's activation, rather than as minimal dependency sets.
More expressive relevance measures or input-conditional selection strategies may further reduce the number of required components without sacrificing fidelity.

The first Transformer layer constitutes a notable exception: fidelity degrades only mildly even when as little as $1\%$ of upstream components are retained.
In this layer, FFN pre-activations appear to be dominated by contributions from token (and positional) embeddings, with comparatively limited influence from early attention outputs.

Importantly, the number of upstream components that actively influence a neuron is even smaller than the nominal sparsity levels suggest. For a given input, only a subset of neurons and attention dimensions in each layer are active.
For example, at a sparsity level of $10\%$, if approximately $5\%$ of neurons in a layer are active, then only about $0.5\%$ of all possible neuron outputs contribute meaningfully to the activation of a given neuron.
This compounding effect underscores the highly selective nature of inter-layer communication in FFNs and suggests that neural dependencies in Transformers are both sparse and structured.

Figure~\ref{fig:layerDensityPlot} visualizes where the relevant upstream components originate, showing for each target layer the fraction of neurons and attention dimensions selected from each preceding layer at a sparsity level of 15\%.
Several patterns emerge. First, we observe locality: layers immediately preceding the target layer typically contribute a disproportionately large share of the relevant components, suggesting that adjacent computations exert stronger influence on downstream neurons.
Second, we observe layer-wise variability: certain layers consistently contribute more relevant components than others, with distinct patterns for FFN neurons and attention outputs.
In the GPT-2 family, layers in the upper half of the model tend to play a broader role, whereas lower layers are more localized.
This suggests that some layers serve as broader information hubs, while others play more specialized or localized roles.
For Qwen2.5-0.5B, subsequent MLP blocks generally place greater emphasis on attention layers, particularly in the early and middle parts of the model.
Another noteworthy observation is that in some models, including GPT-2 M, GPT-2 XL, and Qwen2.5-0.5B, the first attention layer does not appear to contribute to the computation of subsequent FFN neurons, as almost none of its components are selected.
This complements the earlier finding that the first FFN layer is dominated by token and positional embeddings.

\subsection{Model-Level Intervention}
\label{sec:perplexity}

\begin{figure*}[bt]
  \begin{center}
    \centerline{\includegraphics[width=\linewidth]{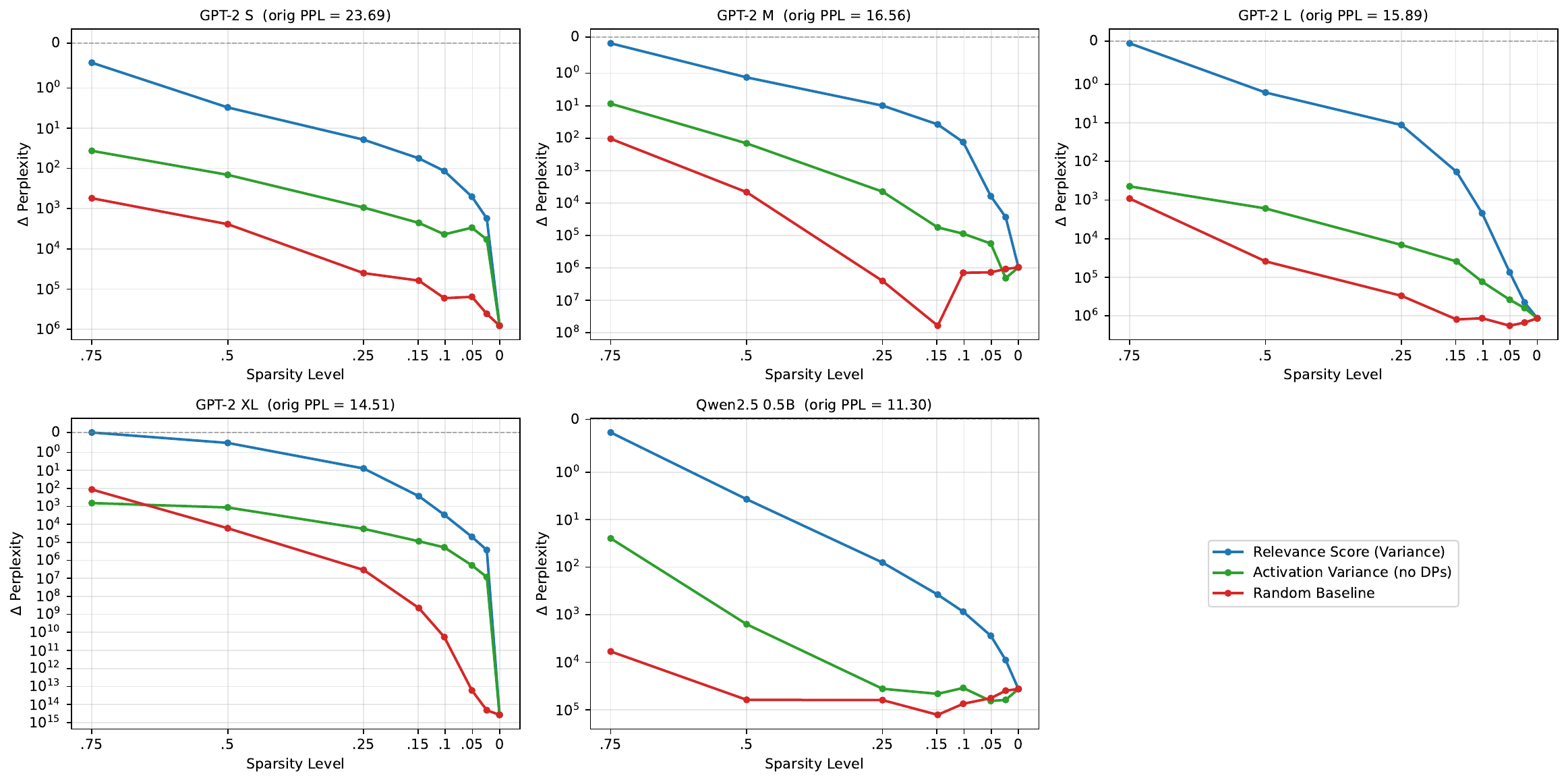}}
    \caption{
Perplexity delta on the MiniPile test subset when the neuron-specific masks are applied simultaneously in all layers, plotted against the sparsity level. Each panel corresponds to one model; curves compare our relevance scoring, an activation-variance-only ranking corresponding to a network pruning approach, and a random selection baseline.
}
    \label{fig:perplexityPlot}
  \end{center}
\end{figure*}

The preceding analysis evaluates each target neuron in isolation.
To assess whether the identified sparse dependencies support the computation of the model as a whole, we apply the neuron-specific masks in all FFN blocks across all layers simultaneously and measure the resulting perplexity on the disjoint MiniPile test subset under tightening sparsity levels.
We compare three selection modes.
In the first, components are selected per target neuron according to our relevance scores.
The second mode ranks components solely by the variance of their activations, $\mathrm{Var}(\hat{c}_s)$, independent of any target neuron; this corresponds to a network pruning baseline~\citep{DBLP:journals/corr/HanMD15} in which only the neurons and attention dimensions with the highest activation variance are retained and the rest are pruned, i.e., mean-ablated.
We also evaluate this target-independent ranking at the neuron level in our ablation studies (Figure~\ref{fig:aucAblations}).
The third mode selects components uniformly at random.

This is a challenging setup: mean-ablation inevitably introduces small deviations in every patched pre-activation, and with the intervention active in every layer, these errors multiply across depth.
The perplexity gap to the unmodified model is therefore expected to widen as sparsity tightens.
The more informative signal is where this degradation accelerates and the distance to the activation-variance-only curve narrows, marking the sparsity regime in which targeted, neuron-specific selection loses its advantage over target-independent pruning.
As shown in Figure~\ref{fig:perplexityPlot}, this transition occurs between $10\%$ and $5\%$ sparsity for GPT-2 S and GPT-2 M, around $15\%$ for GPT-2 L and GPT-2 XL, and between $25\%$ and $15\%$ for Qwen2.5-0.5B.
For Qwen2.5-0.5B, the intervention is particularly demanding: owing to the gated FFN design, each logical neuron combines the nonlinear gating activation with the output of a parallel linear transformation of the residual stream (cf.~Appendix~\ref{sec:adaptations}), and we ablate the upstream contributions of both pathways.

Also of note is the contrast at moderate sparsity: with random selection, masking only $25\%$ of the incoming connections on average already leads to a notable degradation, whereas relevance-based selection leaves perplexity barely changed, despite the numerical deviations that the intervention necessarily introduces in every layer.

\subsection{Ablation Studies}

\begin{figure*}[hbt]
  \begin{center}
    \centerline{\includegraphics[width=\linewidth]{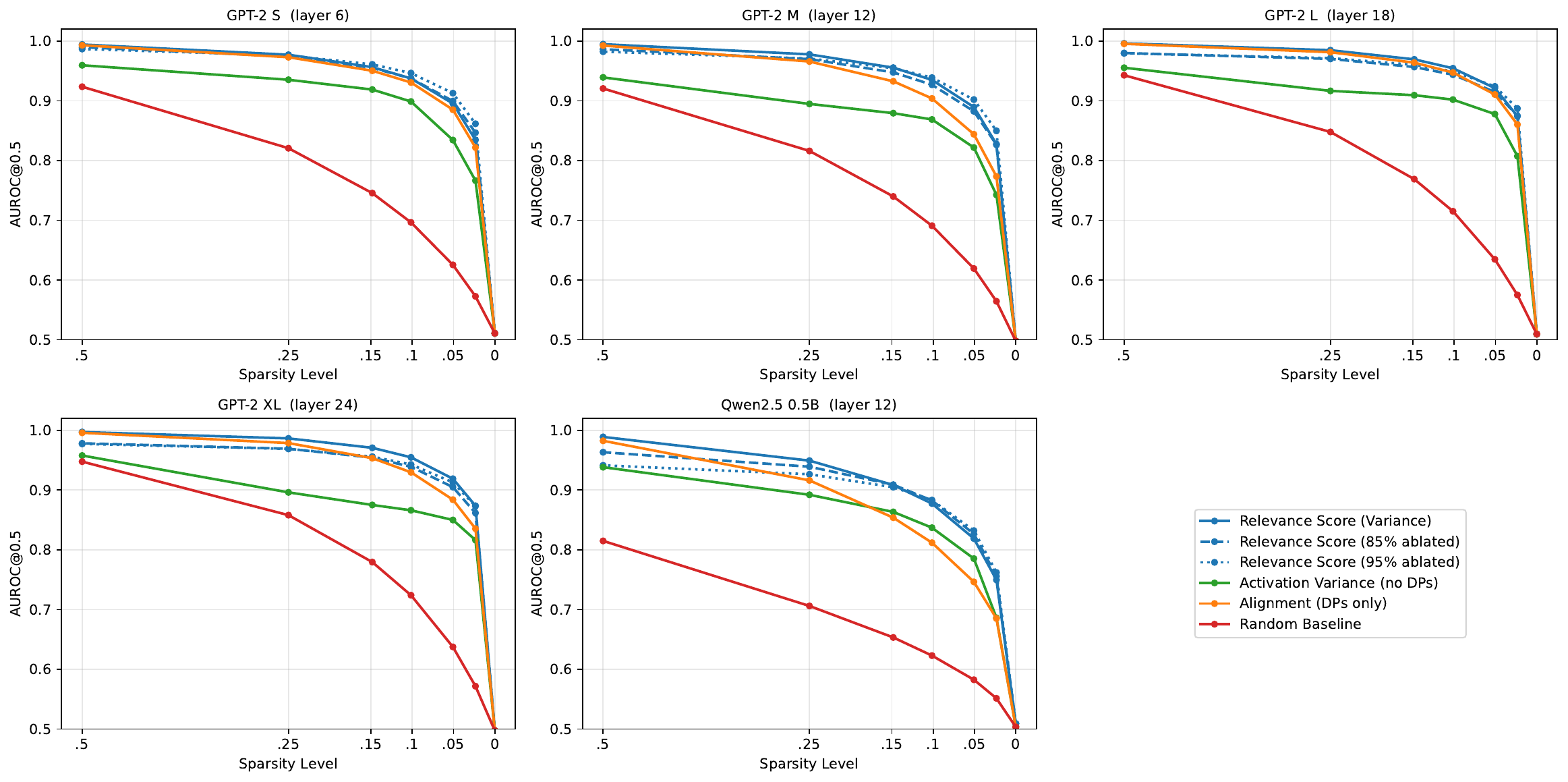}}
\caption{
Ablation results for the middle layer of each model, showing the AUROC for predicting meaningfully active FFN neurons from patched pre-activations. We compare the full relevance score against rankings based on weight alignment only (dot products), activation variance only, and random selection, as well as two runs of the full score in which upstream activations below the 85th and 95th percentile of each layer's activation distribution are ablated.
}
    \label{fig:aucAblations}
  \end{center}
\end{figure*}

We conduct ablation studies to disentangle the contributions of the two factors of our relevance score, weight alignment and activation variance, and to quantify the compound sparsity effect described in Section~\ref{sec:results}.
Figure~\ref{fig:aucAblations} reports AUROC scores for the middle layer of each model.

\textbf{Contribution of the scoring components.}
To isolate the two factors of the relevance score, we compare the full method against baselines that rank upstream components by weight alignment alone, i.e., the magnitude of the dot product between neuron input weights and upstream output vectors, by activation variance alone, i.e., the target-independent, pruning-style ranking of Section~\ref{sec:perplexity}, and by uniform random selection.
Weight alignment alone already yields strong selections for several models, with AUROC scores close to the full relevance score for GPT-2 S, GPT-2 L, and GPT-2 XL.
In contrast, GPT-2 M and Qwen2.5-0.5B benefit visibly from incorporating activation statistics, particularly at lower sparsity levels.
This suggests that the empirical variances of upstream activations are spread less uniformly in these models: where components differ little in how strongly their normalized activations fluctuate, geometric alignment alone determines the ranking, whereas a broader spread of activation variances makes the activation statistics informative.
The variance-only ranking achieves markedly lower fidelity than the full relevance score throughout, confirming that geometric alignment carries the bulk of the target-specific signal.
Random selection, finally, degrades quickly even at modest sparsity levels, with AUROC scores dropping to roughly $0.7$--$0.86$ across models when only $25\%$ of upstream components are retained, underscoring that the fidelity achieved by relevance-based selection reflects genuinely selective dependencies rather than mere redundancy.

\textbf{Compound sparsity under upstream activation masking.}
To quantify the compound sparsity effect, we evaluate the full relevance score under constrained upstream activation.
Specifically, activations from all preceding layers are masked to their average non-firing value whenever they fall below the 85th or 95th percentile of the layer-wise activation distribution, retaining contributions only from the top 15\% or top 5\% of neurons, respectively.
Ground-truth target activations are computed from the unmodified inputs.

Despite these severe constraints, the resulting curves nearly coincide with the unconstrained relevance score.
This confirms that of the meaningfully active upstream neurons, only a fraction is relevant for a particular target neuron: the effective fan-in (the number of upstream components that are both active and influential) is substantially smaller than the nominal subset size $K$ selected by our relevance scoring.
Hence, activation sparsity in upstream layers compounds with dependency sparsity, yielding highly selective inter-layer communication in practice.

\subsection{Limitations}
Our study provides an empirical, descriptive analysis of sparse inter-layer dependencies in FFN neurons. The following limitations clarify the scope and interpretation of our results.

First, our relevance scores identify subsets of upstream components that are sufficient to preserve FFN neuron activations under mean-ablation of the remaining components, but they do not guarantee minimal or optimal dependency sets. Reported sparsity levels should therefore be interpreted as upper bounds on functional dependency size. More compact supports may be recoverable using alternative or input-conditional relevance criteria.

Second, the mean-ablation intervention in our neuron-level analysis applies only to the direct contributions entering the target neuron; the induced deviations are not propagated through intermediate layers. The model-level intervention of Section~\ref{sec:perplexity} complements this local analysis by activating the neuron-specific masks in all layers simultaneously, so that deviations carry through the entire network and compound across depth, and the modest perplexity degradation at moderate sparsity provides propagated, model-wide evidence for the identified dependencies. However, perplexity is an aggregate measure and may mask behavioral changes on specific tasks or inputs. Moreover, both analyses establish the sufficiency of sparse pathways rather than their causal necessity.

Third, our evaluation covers models of up to 1.5B parameters and relies on a single, albeit diverse, corpus for estimating activation statistics and evaluating fidelity. Whether the observed degree of dependency sparsity persists in substantially larger or instruction-tuned models remains to be verified.

Finally, although our findings suggest opportunities for inference optimizations that exploit sparse neuron-to-neuron interactions, realizing such improvements would require additional systems-level and hardware-aware development.

\section{Conclusion}

We investigated whether the dense parameterization of Transformer FFN blocks conceals sparse functional dependencies at the neuron level.
To this end, we developed a training-free attribution method that decomposes neuron pre-activations into contributions from upstream neurons and attention outputs, combining weight-based alignment with empirical activation statistics.
A deliberate design choice was to rely as directly as possible on the model's native weights, avoiding learned feature extractors and requiring only efficiently computable activation statistics.
Our experiments across GPT-2 and Qwen2.5 demonstrate that FFN neuron activations are faithfully preserved when all but a small subset of upstream contributions are replaced by their average values, with effective sparsity compounding when accounting for the inherent activation sparsity of preceding layers.

These findings carry several implications.
First, the observation that neurons depend on sparse, structured subsets of earlier computations suggests that information flow in Transformers is more selective than dense weight matrices would imply, potentially informing the design of sparse-by-construction architectures.
Second, by identifying candidate sparse pathways, our method provides a practical tool for circuit-level interpretability, complementing fine-grained causal tracing and feature-based decomposition.
Third, the identified dependency structure could guide model distillation: pruning components outside the sparse support for specific tasks may yield compact, efficient architectures without full retraining.
Finally, the success of our decomposition, which treats neuron output vectors as directional units of contribution to the residual stream, provides empirical support for the linear representation hypothesis, corroborating the view that Transformers operate via linear superposition of directional features.

Several directions remain for future work: extending the analysis to attention heads, disentangling how attention aggregates information across positions before writing to the residual stream, and leveraging sparse dependency structure for practical efficiency gains through specialized computation kernels or conditional execution.

\bibliography{references_clean}
\bibliographystyle{icml2026}

\newpage
\appendix
\onecolumn

\section{Adaptations for Recent Architectures}
\label{sec:adaptations}

For clarity and conciseness, the main text focuses on a standard GPT-style Transformer architecture. However, many recent models such as Llama and Qwen introduce several architectural modifications that require minor adaptations of our analysis.

First, these models commonly employ rotary positional embeddings applied within each attention block, rather than adding absolute positional embeddings to the residual stream. This change does not affect our method, as positional information is already incorporated into the attention outputs that enter the residual stream.

Second, LayerNorm is often replaced with RMSNorm, which omits centering and normalizes activations using only the root-mean-square statistic. Our decomposition readily extends to RMSNorm by adjusting the normalization term accordingly.

The most significant difference concerns the FFN activation function. Modern architectures frequently replace standard elementwise nonlinearities with gated mechanisms, in which a nonlinear gating function is multiplied by a second term obtained via a linear projection of the residual stream.

For the layer-wise analysis in Section~\ref{sec:results}, we focus on the activation of the gating component itself, treating it as the effective nonlinear neuron. This enables direct comparison with standard ReLU/GELU neurons across architectures and reflects the role of the gating activation as the primary source of nonlinearity. For determining the relevance of upstream components, we use the combined output of the neuron (the gating activation multiplied by the corresponding linear projection) as the effective scalar activation, since it is this product that scales the associated output vector added to the residual stream.

For the model-level intervention of Section~\ref{sec:perplexity}, however, we additionally apply the same relevance-based sparsity intervention to the linear branch of the gated FFN. Specifically, for each neuron we also mask the upstream contributions to the parallel linear projection, retaining only the top-$K$ components by relevance score. This ensures that both multiplicative factors of the gated output are subject to the sparsity constraint, reflecting the full impact of the intervention.

\end{document}